\documentclass[12pt]{article}
\usepackage{graphicx}
\usepackage{amsfonts}
\usepackage{latexsym}
\usepackage{amsmath}
\makeatletter
\@addtoreset{figure}{section}
\def\thefigure{\thesection.\@arabic\c@figure}
\@addtoreset{table}{section}
\def\thetable{\thesection.\@arabic\c@table}

\def\@sect#1#2#3#4#5#6[#7]#8{\ifnum #2>\c@secnumdepth
     \def\@svsec{}\else
     \refstepcounter{#1}\edef\@svsec{\csname the#1\endcsname.\hskip .75em
}\fi
     \@tempskipa #5\relax
      \ifdim \@tempskipa>\z@
        \begingroup #6\relax
          \@hangfrom{\hskip #3\relax\@svsec}{\interlinepenalty \@M #8\par}%
        \endgroup
       \csname #1mark\endcsname{#7}\addcontentsline
         {toc}{#1}{\ifnum #2>\c@secnumdepth \else
                      \protect\numberline{\csname the#1\endcsname}\fi
                    #7}\else
        \def\@svsechd{#6\hskip #3\@svsec #8\csname #1mark\endcsname
                      {#7}\addcontentsline
                           {toc}{#1}{\ifnum #2>\c@secnumdepth \else
                             \protect\numberline{\csname the#1\endcsname}\fi
                       #7}}\fi
     \@xsect{#5}}
\def\@begintheorem#1#2{\it \trivlist \item[\hskip \labelsep{\bf #1\ #2.}]}
\def\section{\@startsection {section}{1}{\z@}{-3.5ex plus -1ex minus
 -.2ex}{2.3ex plus .2ex}{\normalsize\bf}}

\begin{document}

\title{Theory of sexes by Geodakian as it is advanced by Iskrin}
%uncomment next line to remove date
\date{} 
\maketitle

\begin{center}
%\small
\author{
Boris Lubachevsky\\
{\em bdl@bell-labs.com}\\
Bell Laboratories\\
600 Mountain Avenue\\
Murray Hills, NJ 07974\\
}
\end{center}

\setlength{\baselineskip}{0.995\baselineskip}
\normalsize
\vspace{0.5\baselineskip}
\vspace{1.5\baselineskip}
%\end{center}

% ----------------------------------------------------------------
\begin{abstract}
In 1960s V.Geodakian proposed a theory that explains sexes
as a mechanism for evolutionary adaptation of the species
to changing environmental conditions. 
In 2001 V.Iskrin refined and augmented the concepts of Geodakian
and gave a new and interesting explanation to several phenomena
which involve sex, and sex ratio, including the war-years phenomena.
He also introduced a new concept of the "catastrophic sex ratio."
This note is an attempt to digest technical aspects
of the new ideas by Iskrin.
\end{abstract}
% ----------------------------------------------------------------

\section{Introduction}
\hspace*{\parindent}
In several publications which date back to 1960s
cyberneticist Vigen Geodakian 
puts forward a theory that explains sexes.
(See his site {\em geodakian.com}; 
an easy-to-read article "Why two sexes" by Geodakian
is web-available in English, see 
{\em arxiv.org/abs/cs.NE/0408006})
According to the theory, the sexual mechanism of reproduction
provides evolutionary advantages for the species that employs it. 
The 2001 book "Dialektika polov" by Vladimir Iskrin
presents Geodakian's concepts substantially refined and augmented.
Iskrin also publishes an article "Catastrophic sex ratio" devoted to 
his new concepts in the subject.
(Both the book, in Russian, and the article, in English, 
can be found in {\em iskrin.narod.ru}.)

In this note I discuss the contributions made by Iskrin
to the theory of sexes.
This is not a review of either the book or the article.
I omit many interesting observations
in history, psychology, philosophy, and even politics, made by Iskrin
who views his writings in general as a philosophy.
The present note is an attempt 
to digest the mere technical aspects of Iskrin's ideas.
The note 
has not been authorized nor approved by either
of the two authors and 
I do not guarantee that my interpretation 
fully agrees with the originals.
In case of any doubt 
an interested reader is referred to the sources indicated above.

\section{Life expectancy and size}
\hspace*{\parindent}
While agreeing with Geodakian on that 
the sexual reproduction is evolutionary more advantageous
than the asexual one,
Iskrin also pinpoints a specific reason 
for the emergence of the sexes.
The reason is basic, it exists across different species.
Namely, we, the complex multi-cellular organisms,
better be sexual because we live long.  
Long, that is, by comparison with the simple organisms.
Longer the life expectancy of a species' members,
more evolutionary pressure is exerted
on the species to become sexual.

It can be added that being multi-cellular our large size as compared
with the size of simple organisms also contributes to the pressure.
Indeed, the number of microbes of a particular species
thriving under appropriate conditions (temperature,
availability of nutrients, etc) in a small pool of water 
might be comparable with the entire human population.

For single-cell or other simple organisms, 
the life duration is the time elapsed from the moment 
of setting the new organism's genetic makeup 
to the moment when the organism is able to divide or otherwise reproduce
and to set a genetic makeup of its offsprings.
The life duration is understood 
in a usual way for complex multi-cellular creatures: 
it is the time from the birth to the death.

Here is the argument.
The life expectancy of individuals in a species 
divided by their number
is inversely proportional
to the species' renewal rate. 
(The renewal rate is the rate at which new organisms
substitute the old ones.
It is the number of new organisms produced per unit of time,
if the population size remains stationary.
The ratio of the renewal rate of a microbe species 
over that of humans
might easily be greater than ten orders of magnitude.)
On one hand, for an asexual species to survive, its
renewal rate has to be high enough for the species' genotype to be
able to track and adapt to the environmental change.
The tracking occurs by the way of a simple random search
whereby
the species uses the Darvinian survival-of-the-fittest selection
superimposed on the random mutations. 
And those mutations occur with the rate
proportional to the renewal rate.
On the other hand, 
the sexual reproduction mechanism compensates the species'
slowness, if any, in renewal.
That is, the sexual species relies not only on 
the simple random search via mutations
at renewals
for tracking the environmental change, but also and primarily
on the sexual selection.
The mechanism of sexual selection is more complex than the basic
random search. Its description
constitutes 
the Geodakian's base theory of sexes, 
and I skip here the explanation, 
see, e.g., 
{\em arxiv.org/abs/cs.NE/0408006}.

A species of living long and large-sized individuals if
the sexual reproduction 
is {\em not} employed
is more likely to become extinct
than one that {\em does} employ it.
This holds assuming
that all the other conditions between the two cases
are the same and that the environment changes fast.

It may be added that while the longer life and larger size
of complex organisms
puts the species at a disadvantage as  the
mechanism of evolutionary adaptation
of the species to changing environment is concerned,
the complexity itself offers many advantages.
The complexity is an evolutionary result of accumulating mechanisms 
that serve to withstand a possibly larger variety of environments.
A complex creature withstands the environmental change better
than a simple creature as long as the changed condition
is within the band of acceptable conditions.
For a complex creature such band is wider than for a simple creature.
This is a "resourceful" method: for different conditions the creature
offers different mechanisms. The body complexity and size are utilized
to host these different mechanisms.

Occasionally,
the nature 
offers what seems like counter-examples: 
populations of 
long-living large-sized multi-cellular organisms,
which none-the-less are asexual.
According to Iskrin these examples do not contradict the rule.
Such populations 
were lucky enough to enter exceptionally
stable and comfortable environments.
Another way of saying this is that
the genotype of such species has accumulated and presently holds
all the mechanisms and features that are
necessary to prosper in the current environment
and in its  possible variations.
The sexual reproduction mechanism of such populations
has since
become redundant, and it has weakened and/or disappeared.
These populations usually include only females
and in the absence of males they reproduce parthenogenetically.

A particular attention is paid to
the life duration of an individual as an operating
factor in the mechanism of sexual reproduction.
This distinguishes 
the treatment of the problem by Iskrin from that by Geodakian. 
The life duration is involved in various ways
in all the proposed new mechanisms and theory refinements.

\section{Sex ratio}
\hspace*{\parindent}
For example, Iskrin considers the sex ratio ($SR$), a
much studied statistics for the humans and other sexual species.
This is the number of the living males 
divided by the number of the living females. 
The statistics is usually reported as being multiplied by 100.
For instance, 
a typical reported sex ratio at birth for humans
is $SR=105$ which means that among the newborns
on average there are 105 boys for 100 girls.
The life duration enters 
the consideration of the statistics
by segregating the $SR$ according to the individuals' age 
so that to each age interval its own $SR$ corresponds.
If the granularity of the segregation is made very fine,
then, at least ideally, the $SR$ can be thought of
as a continuous function of 
the continuous time-like age paremeter $t$, 
let us denote this function $SR(t)$.

In agreement with the base theory of sexes,
in any age group $t$,
the chance of a death for a male is greater
than that for a female.
(This is usually confirmed by statistics, 
but Iskrin also notes
distortions of this rule in certain
human subpopulations in the former USSR.)
Therefore $SR(t)$ monotonically
decreases with $t$: $SR(t_1) > SR(t_2)$ if $t_1 < t_2$.

The aging begins at $t=0$,
the moment of conception. 
The sex of an embryo is set at conception, 
so the values of $SR(t)$ make sense for $0 \le t \le t_b$,
where $t_b$ is the time elapsed  from conception to birth.
(Among many implied idealizations of the model, $t_b$ is assumed
to be the same for all the individuals in a population.)
For humans, various estimates
give values 
from $SR(0)$=120 to $SR(0)$=180.
$SR(t)$ gradually reduces to 0 for large enough $t$,
because women live longer than men do, on average.

\section{Parity age}
\hspace*{\parindent}
It is proposed to view
the sex ratio profile, 
especially its rate of descent, $\frac {d} {dt} SR(t)$,
as a measure of the harshness of the environmental conditions
and/or of the speed of their change.
The {\em parity age} $t_p$ is introduced.
By definition, at $t = t_p$ the profile
crosses the value of 100.
Thus at the parity age 
the number of living
males is equal to the number of living females.
Because of the monotonicity and  continuity of the profile, 
and because of its
starting value $SR(0)$ being greater than 100 
and ending value $SR($large $t)$ being close to 0,
parameter $t_p$ is well-defined.
Assuming tacitly
that the entire profile of the functions $SR()$ is determined
by this single parameter,
the $t_p$ now measures the harshness or the speed of change of the 
environment:
smaller the $t_p$ faster the descent of $SR(t)$ and harsher
or changing faster the environment is toward the population.

\section{Quality of an individual}
\hspace*{\parindent}
Along the footsteps of Geodakian, Iskrin considers
quality, quantity, and assortment (that is, variety or diversity)
as the main operating parameters in the sexual mode of reproduction.
He then concentrates
on the parameter of quality of an individual ($Q$).
The $Q$ measures the degree of adaptability of the individual
to the current conditions offered by the environment, 
the degree of the ability to adjust to and be
comfortable in the environment.

By contrast with the unambiguous
definition of the parameter $SR$ as above,
the given definition of the parameter $Q$ is vague.
It is open to 
different interpretations let alone that no method
of assigning a concrete
numeric value to $Q$ is proposed by Iskrin
(or by Geodakian, for that matter).

It seems to me that there may be
at least three different directions of making 
the definition of $Q$ more precise:

I) "genetic" $Q = Q^g(t) = const$, 
the $Q$ stays {\em unchanged} over the lifetime
of the individual because it is determined entirely by the
genome;

II) "aging" $Q = Q^a(t)$ which monotonically {\em decreases} over time,
as the individual ages while exhausting the living resource
used to withstand the environment;

III) "wisdom" $Q = Q^w(t)$ which monotonically {\em increases} over time,
as the individual matures while learning how to better adapt to the
environment.

These three directions for defining the $Q$, 
differ primarily in the behavior of $Q=Q(t)$
with respect to the age
parameter $t$.
The parameter
is the "speciality domain" for Iskrin.
By contrast, 
for the "lump" version of the $Q$ in Geodakian's
treatment,
without segregating according to $t$, the question
of these distinctions does not arise.

Perhaps, a better model would be some combination of the three:
$Q(t) = f[ Q^g(t), Q^a (t), Q^w (t) ]$ where function $f$ is 
not necessarily linear.
However, as the subject is further discussed in the book,
it appears that Iskrin adopts the simplest 
it-is-all-in-the-genes definition so that
$Q (t)$ remains constant throughout the life of an indifidual. 

So $Q$ is not well-defined. But {\em if} it were
defined well, 
one could operate with the $Q$ measure
in the same
manner as with the $SR$ measure.
Namely, a statistically averaged $Q$ could be considered and
the average could be segregated by the age.
Furthermore, the average could be segregated by the sex.
The average quality of males $Q_m(t)$
could be compared with that of females $Q_f(t)$ at different ages $t$.
Iskrin is doing just that
in his book, ignoring the fact that the $Q$ is underdefined.

\section{Parity in quality}
\hspace*{\parindent}
As mentioned above,
Geodakian does not segregate anything according to the life duration,
in particular, he does not segregate the $Q$.
He makes a "lump" statement to the effect that
the average quality of a male is lower than that of a female.

But Iskrin does segregate.
According to him, 
the average quality of a living individual 
in the population, $Q(t)$, increases with $t$.
This is so,
because lower $Q$ individuals 
remove themselves from the averaging sooner
than the higher $Q$ individuals.
Note that the increase would take place if the quality is
"genetic" $Q=Q^g(t)$.
The latter is apparently the assumption by Iskrin.
And the average "wisdom" quality would also increase
but not necessarily the average "aging" quality.

Furthermore, 
the average $Q_m(t)$ increases faster, 
than the average $Q_f(t)$.
This is so, because the weaker sex 
(those are the males, see 
{\em arxiv.org/abs/cs.NE/0408006}
)
depletes its lower $Q$ members at a faster rate 
than the stronger sex.

At $t=0$ (conception), 
$Q_m(t) < Q_f(t)$.
This is in agreement with Geodakian.
But as 
the lower $Q$ individuals die out for $t > 0$
the ratio $Q_m(t)/Q_f(t)$ monotonically increases
while inequality $Q_m(t) < Q_f(t)$ still holds.
For a sufficiently large $t > t_p$
the inequality between $Q_m(t)$ and $Q_f(t)$ reverses
sense and becomes $Q_m(t) > Q_f(t)$. 
Here $t_p$ is the 
age of parity in the average quality between the sexes:
$Q_m(t_p) = Q_f(t_p)$.

Note that the age of parity in quality needs not be the same
as the age of parity in numbers of members of the two sexes, 
discussed in Section 4 above.
A sort of a "proof" is suggested that for humans these
two ages better be the same or differing only slightly.
Iskrin's "proof" appeals to an argument of the economy of "material." 
The argument is that the evolution benefits a population
that avoids an unnecessary waste of its members.

Iskrin assumes that the mating occurs in pairs
of the same age on average and of the same quality on average
and therefore if the numbers were not equal there would be a waste.
However, Iskrin also suggests elsewhere in the book,
that, on average, there is,
perhaps, a small systematic discrepancy in quality
between a pair of mating individuals, namely, an
increase from the female to the male. Furthermore,
there exists a similar systematic discrepancy in the ages
between a pair, also an increase from the female to the male.
Because of these discrepancies,
a small discrepancy may exist, he concedes,
between the two parity ages.

If the argument above seems vague to a reader-mathematician,
so it does to me. That is why I enclose "proof" in quotes.
Perhaps, the theory can be mathematically completed in such a
way that the equality or small discrepancy
between the two parity ages would indeed follow
from the definitions as a simple theorem.

Among the
statistically observed phenomena that confirm
the changing relation between $Q_m(t)$ and $Q_f(t)$
is the one that says that, on average,
a teenage boy-driver is more prone 
to an automobile accident than a teenage girl-driver,
whereas for senior drivers the relation is reverse.

Iskrin also cites statistics which indicates 
that in the 1950s, in the after-the-WWII USSR, 
the humans' $t_p$ was smaller than 20 years
and then, as the conditions improved, 
the $t_p$ gradually increased to reach in 1980s almost 30 years,
the value, he thinks, the $t_p$ should be equal to
in a comfortable social environment.
In pre-civilization environments, he believes, the humans' $t_p$
was 12-14 years.

\section{Which sex is more responsible for forming the sex ratio}
\hspace*{\parindent}
The base sex theory of Geodakian states that 
faster the environment is changing, 
larger the target $SR$ is being set by a sexual population so that
more new males are being produced 
in proportion to the
number of the new females.
But where in the species is the mechanism of adjusting the $SR$
located and how specifically does it operate?
Which sex in the population is more responsible for forming the $SR$?

Iskrin furthers the theory by attempting to answer
ohese questions.
Implications of 
his answers can be verified
statistically.
In my opinion, 
the answers allow one to explain 
convincingly
certain existing statistical and natural phenomena.
These will be discussed below.

Iskrin argues that in a sexual species 
the heterogametic sex is primarily responsible
for forming the $SR$.
(The heterogametic sex is the one that supplies two sorts
of gamets, X and Y. The opposite sex is homogametic,
it produces only one sort or gamets, X.)
Among the mammals, particularly, for the humans,
the heterogametic sex is the males. 

By Iskrin,
each individual man controls the $SR$ in his progeny
by setting the pre-conception $SR$ in his sperm.
Therefore, the controlling parameter is the ratio
of the number of Y spermies over the number of X spermies. 
Iskrin stops short of saying that this value
is, in fact, equal to the mathematical expectation of
the sex ratio at conception $SR(0)$ 
(for the individual father, not the average over the population).
There might exist some 
hormonal and other biological "filters"
in the woman's body which change the pre-conception
sex ratio.

The existence of such filters is strongly suggested
by the existence
of some family trees where only girls were
born in each generation and where
this property was inherited through the mother.
However, the apparent strength of the filters for such
rare families is not a rule for all humans.
On average,
the effect of such filters 
on the deviation of $SR(0)$ from its parity value $SR(0)=100$
is much weaker than the effect of the initial setting 
which is formed exclusively in the man's body.

Continuing this discourse mainly for the case of humans,
Iskrin next proposes that the man, his body, and especially,
his sex organs is sort of a "sensor" in the mechanism
of controlling the man's individual expected $SR$. 
He makes an interesting observation
about the natural built-in sensitivity and protection
of the sex organs in the body.
Man's sex organs are very sensitive 
and yet they are not well protected,
being positioned outside of the protective body bulk.
Such positioning of the man's sex organs
is obviously against the private
interest of the individual man, 
but it is expedient for the population as a whole.
It allows the male's sex organs 
to perform better their sensory function,
for instance, to sense the outside temperature.
The safety of the male's sex organs
is sacrificed 
for the same cause the life itself of
an individual male, if he is of a lower quality, unfit,  
is sacrificed. 
This cause is speeding up the adaptation of the species
to the changing environment.

The proposed parameter which controls the $SR$,
namely the 
pre-conception $SR$ in the man's sperm,
fits well with statistically observed phenomena and this fit
will be discussed below. But before that, I should explain
an item in Iskrin's theory which is completely new, as compared
with Geodakian's theory.

\section{Iskrin's "catastrophic sex ratio"}
A population of an initially sexual species may 
become
parthenogenetic (all-female) if
it finds itself in an
exceptionally comfortable and stable environment.
In such an environment
the population sets a target 
"comfortable sex ratio" $SR = 0$.

Iskrin also considers a diametrically opposite scenario.
In it a sexual population encounters conditions
which are interpreted by its members as exceptionally bad and unstable.
In such a state, the population "feels" threatened
with what is perceived as an imminent extinction,
a catastrophe.
In this emergency state a special target $SR$ is set. 
Paradoxically, 
as a number, this "catastrophic sex ratio" 
is the same as the "comfortable sex ratio" $SR = 0$.

This paradox is 
illustrated with a metaphor.
The population is compared with a canoe rider.
The sexual mechanism of reproduction is compared
with the steering paddle.
If the flow is laminar and very steady the rider 
may not even need to use the paddle. 
On the other hand, if an extremely turbulent flow overturns
the canoe the rider would not need to use the paddle either.

Iskrin cites several examples that involve the catastrophic $SR$,
including examples of sexual plants.
Some
plants change their sex in response to their worsening conditions.
When the conditions become mildly bad, female plants may become males.
This exemplifies the mechanism of increasing the $SR$. 
When the conditions become very bad, male plants may become females.
This exemplifies the mechanism of setting the catastrophic $SR$.
The gardeners, farmers or other plant care takers
systematically "condition" the plants
(cut brunches, flowers, etc.)
to achieve desired plant properties,
for instance, the ability to produce tasty fruits.
Such "conditioning" is interpreted by the plants
as none else but mutilation, abuse.
It awakens the evolutionary developed response of the sex change,
which is a defence of the species.

Also, 
a research paper is cited which 
observes an unusually low $SR(t_b)$, less than 40,
in the families where the father's occupation,
such as a geologist, a high-altitude scientist,
makes him to stay up in the mountains for long
periods of time.
The low $SR(t_b)$ has nothing to do with 
the air, water, difference in heights
and other such physical factors conjectured as the cause
by the author of the paper.
Instead, Iskrin sees in this case a clear involvement of
the catastrophic $SR$ phenomenon. 
More specifically, 
it is the result of a prolonged abstinence.

For many millions of years of our animal prehistory
the primary 
reason for male's extended abstinence was
the absence of females.
The no-female situation 
threatens the species with the extinction.
The animalistic assessment of this fact remains genetically imprinted
in the man's body, who decreases the
pre-conception $SR$ in his sperm when being abstinent.
The mere tens of thousands of years
of our existence as Homo sapiens
have not washed out the mechanism which can be triggered
by the appropriate conditions in
us just as well as in the mice.
This explanation implies
an obvious and 
tangible recommendation for fathers who
wish a son as well as the one for those who wish a daughter.

\hspace*{\parindent}
\section{Why the firstborns are often boys}
\hspace*{\parindent}
Speaking only about humans,
to any person an index value $i$ corresponds, $i = 1,2,3,...$,
such that this person is the $i$th born child of his/her father.
The $SR$ statistics can be segregated by this index.
$SR[1]$ becomes the sex ratio of first-born children,
$SR[2]$ the ratio of the second-born children, etc.
(I denote the segregation by order index $i$ 
with the square brackets, $SR[i]$,
and that differs from the notation for the age segregation; 
the latter
is denoted using parentheses, $SR(t)$.)

The reported average $SR[1]$ is usually noticeably larger than 100.
The subsequent
values of $SR$ are monotonically decreasing:
$SR[1] > SR[2] > SR[3] > ...$.
Iskrin explains the phenomenon by the mentioned in Sec.6
increase of average $Q(t)$ with $t$.
The male's own sensed quality sets the pre-conception $SR$:
a lower $Q$ male sets $SR$ higher, 
and a higher $Q$ male sets it lower.
But the lower $Q$ males also have a higher chance
to remove themselves from the process of reproduction
and that is why the $SR[i]$ decreases with $i$.
$SR[1]$ is higher than 100 because an average young
father has low $Q$.

This phenomenon is compatible with the "genetic" $Q$ model
and even more so with the "wisdom" $Q$ model,
but it might not be compatible with the "aging" $Q$ model.

\section{War years phenomena}
\hspace*{\parindent}
The most familiar war-years phenomenon which involves $SR$
is the increase of $SR(t_b)$ in a nation which participates
in a war.
While a typical peace-time value is $SR(t_b) = 105$,
during a war it usually increases a couple of points and
sometimes it might become as high as $SR(t_b) = 109$.
Such statistically significant increases
persist across different nations and different wars.

In 
{\em arxiv.org/abs/cs.NE/0408006}
Geodakian attributes this increase to a general
tendency of the $SR$ increase under various 
hardships. 
Elsewhere he suggests a more specific mechanism
of the increase as being attributable to
a reduction of the number of men available
for reproduction.

However, Iskrin proposes a different specific mechanism which
might be responsible for the phenomenon.
The phenomenon is caused by a redistribution of $Q$.
A military draftee, man, must pass a medical commission
to become a military serviceman. The commission
filters out physically unfit individuals.
The physical fitness strongly correlates with the quality $Q$,
hence during a war the population of reproductive age men is being
split into the two subsets: those who go fight,
they are, on average, higher $Q$ men;
and those who stay behind, they are, on average,
lower $Q$ men.
The fighting men have substantially smaller opportunity
to become fathers 
than those who stay home.
The non-fighting  lower-$Q$ men
generate progeny of a higher $SR$ 
than the $SR$ would have been 
among all the men capable of reproduction.

A less known war-year phenomenon is a significant
{\em decrease} of the $SR(t_b)$ in a subpopulation which becomes
subject to an extreme hardship, in particular, to starvation.
An example of the phenomenon is 
the 1941-44 blockade by the German troops
of the city of Leningrad 
(presently known as St.Petersburg) in Russia.

The war with Germany
started in the mid-1941, and the German troops fully surrounded
the city of Leningrad by the late fall of 1941.
During 1941 a war-year increase of $SR(t_b)$
was observed in Leningrad. It has reached 106.
However, in 1942 the value of $SR(t_b)$ sharply decreased
to 101 and then again increased to 105 in 1943 and kept increasing in 1944 
to reach 109 in 1945.

It is known that the living conditions in the city, especially,
the food shortages steadily worsened since the beginning of the war
reaching their worst in the winter of 1941-42.
Beginning with the spring of 1942 the conditions,
mainly
the food availability, began to slowly improve
for those who had survived thus far.
The living condition history matches 
the history of the $SR(t_b)$ statistics,
so that 
the 1942 minimum
in $SR(t_b)$ matches the worst food situation in the city
with a several months lag.

Iskrin thinks that a malnutrition 
of mothers during pregnancy
reduces the parity age $t_p$.
The Leningrad blockade starvation sent the $t_p$ value 
to below the $t_b$ value for those
who were born in Leningrad in 1942.
In other words, the male-fetuses were dying 
at an especially high rate as compared with the female-fetuses.
This mother-induced filter was decreasing $SR(t_b)$ while
it was acting in superposition with the father-induced
setting for $SR(0)$. The latter was pushed higher
during the years of the war.
The ambivalent war-years profile of $SR(t_b)$ 
in Leningrad was the result.

During the WWI and the few years after, the Russian cities of
Moscow and Petrograd (another old name for St.Petersburg)
exhibited similarly ambivalent profiles of 
$SR(t_b)$ (high, low, high again, low again) with lows
for years 1919-20 and  then 1924-25.
The lack of food was the main hardship during those "low" years.
The analogous statistics available for the years of WWI 
in the West (Germany, France, England) show
a unimodal increase-decrease profile of $SR(t_b)$ with no ambivalence.

\section{Conclusion}
\hspace*{\parindent}
In my view, Iskrins' theory is interesting
for a researcher in the area in two respects: 
\\
I. It allows one to set experiments and make predictions 
with testable outcomes. 
\\
II. It opens new directions for a mathematical modeling.
  
\end{document}